\documentclass[sigconf,edbt]{acmart-edbt2024}

\def\BibTeX{{\rm B\kern-.05em{\sc i\kern-.025em b}\kern-.08em
    T\kern-.1667em\lower.7ex\hbox{E}\kern-.125emX}}

\usepackage{booktabs} 

\usepackage{multirow}
\usepackage{balance}
\usepackage[normalem]{ulem}
\useunder{\uline}{\ul}{}

\setcopyright{rightsretained}

\acmDOI{}

\acmISBN{978-3-89318-091-2}

\acmConference[EDBT 2024]{27th International Conference on Extending Database Technology (EDBT)}{25th March-28th March, 2024}{Paestum, Italy}
\acmYear{2024}

\begin{document}
\title{WDC Products: A Multi-Dimensional Entity Matching Benchmark}

\author{Ralph Peeters}
\orcid{0000-0003-3174-2616}
\affiliation{%
  \institution{Data and Web Science Group}
  \institution{University of Mannheim}
  \streetaddress{B6, 26}
  \city{Mannheim}
  \country{Germany}
  \postcode{68159}
}
\email{ralph.peeters@uni-mannheim.de}

\author{Reng Chiz Der}
\orcid{0009-0003-8889-3877}
\affiliation{%
  \institution{Data and Web Science Group}
  \institution{University of Mannheim}
  \streetaddress{B6, 26}
  \city{Mannheim}
  \country{Germany}
  \postcode{68159}
}
\email{rder@mail.uni-mannheim.de}

\author{Christian Bizer}
\orcid{0000-0003-2367-0237}
\affiliation{%
  \institution{Data and Web Science Group}
  \institution{University of Mannheim}
  \streetaddress{B6, 26}
  \city{Mannheim}
  \country{Germany}
  \postcode{68159}
}
\email{christian.bizer@uni-mannheim.de}

\renewcommand{\shortauthors}{Peeters et al.}

\begin{abstract}
The difficulty of an entity matching task depends on a combination of multiple factors such as the amount of corner-case pairs, the fraction of entities in the test set that have not been seen during training, and the size of the development set. Current entity matching benchmarks usually represent single points in the space along such dimensions or they provide for the evaluation of matching methods along a single dimension, for instance the amount of training data. This paper presents WDC Products, an entity matching benchmark which provides for the systematic evaluation of matching systems along combinations of three dimensions while relying on real-world data. The three dimensions are (i) amount of corner-cases (ii) generalization to unseen entities, and (iii) development set size (training set plus validation set). Generalization to unseen entities is a dimension not covered by any of the existing English-language benchmarks yet but is crucial for evaluating the robustness of entity matching systems. Instead of learning how to match entity pairs, entity matching can also be formulated as a multi-class classification task that requires the matcher to recognize individual entities. WDC Products is the first benchmark that provides a pair-wise and a multi-class formulation of the same tasks. We evaluate WDC Products using several state-of-the-art matching systems, including Ditto, HierGAT, and R-SupCon. The evaluation shows that all matching systems struggle with unseen entities to varying degrees. It also shows that for entity matching contrastive learning is more training data efficient compared to cross-encoders.
\end{abstract}

%
%



\maketitle

\section{Introduction}
\label{sec:introduction}

Entity matching is the task of discovering records that refer to the same real-world entity in a single or in multiple data sources~\cite{Christen2012DataMC,christophides_end--end_2020}. Entity matching has been researched for over 50 years~\cite{fellegiTheoryRecordLinkage1969}. While early matching systems relied on manually defined matching rules, supervised machine learning methods have become the foundation of most entity matching systems~\cite{christophides_end--end_2020} since the 2000s. This trend was recently reinforced by the success of neural networks~\cite{BarlaugNeural2021} and today most state-of-the-art matching systems rely on deep learning techniques, such as Transformers~\cite{vaswaniAttentionAllYou2017,devlinBERTPretrainingDeep2019,brunnerEntityMatchingTransformer2020,liDeepEntityMatching2020}. 

Over the last decades, a wide range of benchmarks for comparing the performance of entity matching systems has been developed~\cite{primpeliProfilingEntityMatching2020}. Table \ref{tab:comparison} gives an overview of key features of the entity matching benchmarks that are frequently used in recent publications\footnote{\url{https://paperswithcode.com/task/entity-resolution}}. The datasets that are used by these benchmarks range from structured datasets to datasets exhibiting mostly textual entity descriptions. While the older benchmarks focus on matching records between two data sources, more recent benchmarks often rely on data from larger numbers of Web data sources that are more heterogeneous and thus more difficult to match.

While the field of benchmarking entity matching systems is quite mature, existing benchmarks do not explicitly cover the following aspects:

\begin{figure*}
  \centering
  \includegraphics[width=2\columnwidth]{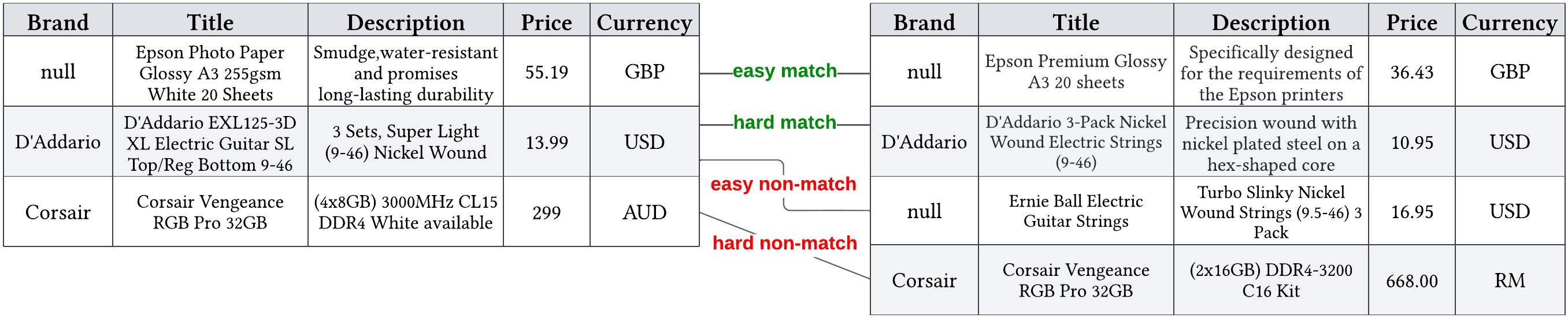}
  \caption{Example of hard and easier matching and non-matching offer pairs from WDC Products. Hard matches (non-matches) exhibit a strong textual dissimilarity (similarity) making them harder to classify correctly.}
  \label{fig:example}
\end{figure*}

\textbf{1. Explicit Modelling of Combinations of Challenges:} The overall difficulty of an entity  matching task depends on a combination of multiple dimensions, such as the amount of hard-to-solve matches and non-matches or the amount of available training examples. Current benchmarks usually represent single points in the space along such dimensions or they provide for the evaluation of matching tools along a single dimension, for instance by offering development sets of different sizes. The benchmarks do not support the systematic evaluation of matching systems along a combination of different dimensions, e.g. offering development sets of different sizes that also exhibit different fractions of corner-cases. Such multi-dimensionality would provide for a more fine-grained evaluation of the strengths and weaknesses of matching systems in a controlled environment.

\textbf{2. Generalization to Unseen Entities:} While state-of-the-art matching systems~\cite{liDeepEntityMatching2020,yaoEntityResolutionHierarchical2022} reach a very high performance for matching entities that are covered by records in the training data, experiments involving unseen entities which are not covered by the training data indicate that the methods lack robustness and perform much worse for such entities~\cite{wangBridgingGapReality2022a,peetersDualobjectiveFinetuningBERT2021}. None of the existing English-language entity matching benchmarks explicitly measures the robustness of matchers with regard to their performance for unseen entities.
Such unseen entities play an important role in many use cases, e.g. new products frequently appear in e-commerce scenarios~\cite{wangBridgingGapReality2022a}, while new books and new articles play a central role in bibliographic data management. 

\textbf{3. Examples per Entity:} Some of the widely used benchmarks only contain a small amount of matching records per entity, e.g. on average around 1.2 records (see Table~\ref{tab:comparison} and discussion in section \ref{sec:adaptionofbenchmarks}). This lack of training records hinders the systematic analysis along the dimension of available training data.

\textbf{4. Entity Matching as Multi-Class Classification:} There are matching use cases that do not require being able to match arbitrary entities but require the matcher to recognize a previously known set of entities. For instance for price tracking or for a market analysis, it might be necessary to find offers for the specific set of products that the company produces. For these use cases, it can be more appropriate to treat entity matching as multi-class classification, e.g. learn how to recognize the relevant products in a large set of product offers.

This paper presents the WDC Products benchmark which addresses the aspects described above. 
It is the first benchmark that employs non-generated real-world data to assess the performance of matching systems along three dimensions: (i) amount of corner-cases, (ii) fraction of unseen entities in the test set, and (iii) development set size. For this, the benchmark provides nine training, nine validation, and nine test sets which can be combined into 27 variants of the benchmark. The range of variants includes simple variants containing only a low amount of corner-cases and no unseen entities while enough training data is available. It also includes highly challenging variants which require the system to match corner-cases involving unseen entities while for the seen entities only a small amount of training examples is available. While these dimensions can be partly supported by existing benchmarks, this usually entails severe limitations especially for modeling the corner-case and robustness dimensions, as we discuss in Section\ref{sec:adaptionofbenchmarks}, which can limit the meaningfulness of the obtainable results.

WDC Products is based on product data that has been extracted in 2020 from 3,259 e-shops that mark up product offers within their HTML pages using the schema.org vocabulary\footnote{\url{https://schema.org/}}. By relying on web data from many sources, WDC Products can offer a decent amount of quite heterogeneous offers per product entity, e.g. WDC Products contains 11,715 product offers describing in total 2,162 product entities belonging to various product categories. Figure \ref{fig:example} shows examples of hard and easy matches and non-matches from the benchmark. 
We test the WDC Products benchmark using the matching systems Ditto~\cite{liDeepEntityMatching2020}, HierGAT~\cite{yaoEntityResolutionHierarchical2022} and R-SupCon~\cite{peetersSupervisedContrastiveLearning2022a} as well as several baseline approaches. The experiments confirm that the benchmark is challenging for state-of-the-art pair-wise matching systems which reach Top-F1 scores between 0.64 and 0.89 depending on the variant of the benchmark. The evaluation also confirms the utility of the 27 variants of the benchmark for the fine-grained evaluation of matchers. For instance, the evaluation along the seen-unseen dimension shows that the performance of all matchers significantly drops for unseen entities.

In order to prevent any potential of information leakage between training and testing, the WDC Products benchmark strictly separates between offers that appear in the pairs which are contained in the training, validation, and test sets, so that each offer can only be contained in exactly one of the splits. This splitting also allows us to offer the benchmark not only in pair-wise format but also as a multi-class matching benchmark while ensuring comparability between the two setups. To the best of our knowledge, WDC Products is the first benchmark to offer fixed datasets for pair-wise and multi-class entity matching tasks, while ensuring comparability and offering multiple dimensions for a systematic evaluation of matchers. 

In summary, the WDC Products benchmark provides the following contributions to the field of benchmarking entity matching systems:
\begin{enumerate}
\item Multi-Dimensionality: WDC Products is the first benchmark to provide fixed training, validation, and test splits along the three dimensions: (i) amount of corner-cases, (ii) unseen entities, and (iii) development set size, resulting in 27 variants of the benchmark for the fine-grained evaluation of matching systems along each dimension in isolation as well as in combination.
\item Generalization to Unseen Entities: WDC Products is the first English-language entity matching benchmark offering the capability to explicitly measure the robustness in the form of generalization performance of matching systems to unseen entities. Besides WDC Products, the Chinese-language benchmark Ember~\cite{wangBridgingGapReality2022a} also provides for measuring this dimension (see Section~\ref{sec:relatedwork}).
\item Pair-wise and Multi-class: WDC Products is the first entity matching benchmark to offer two versions for pair-wise and multi-class entity matching while ensuring comparability between the results.
\item Utility: We use the WDC Products benchmark to compare the recent entity matching systems Ditto~\cite{liDeepEntityMatching2020}, HierGAT~\cite{yaoEntityResolutionHierarchical2022} and R-SupCon~\cite{peetersSupervisedContrastiveLearning2022a}. The evaluation confirms the difficulty of the benchmark and shows that the multi-dimensional design is useful for identifying the strengths and weaknesses of the systems.
\end{enumerate}

The paper is structured as follows: Section \ref{sec:terminology} introduces the dimensions of the benchmark. Section \ref{sec:creation} describes the data collection, data selection, and creation of the benchmark sets. Section \ref{sec:profiling} provides profiling information for all variants of the benchmark. Section \ref{sec:experiments} describes baseline experiments using recent matching systems and discusses the suitability of WDC Products for evaluating these systems. Section \ref{sec:relatedwork} describes related work and finally Section \ref{sec:adaptionofbenchmarks} discusses WDC Products in relation to existing entity matching benchmarks.
The WDC Products benchmark as well as the code used for its generation are available for download\footnote{\url{http://webdatacommons.org/largescaleproductcorpus/wdc-products/}}.

\section{Benchmark Dimensions}
\label{sec:terminology}

This section introduces the dimensions that are used for deriving the variants of the WDC Products benchmark. 
\begin{figure*}
  \centering
  \includegraphics[width=2\columnwidth]{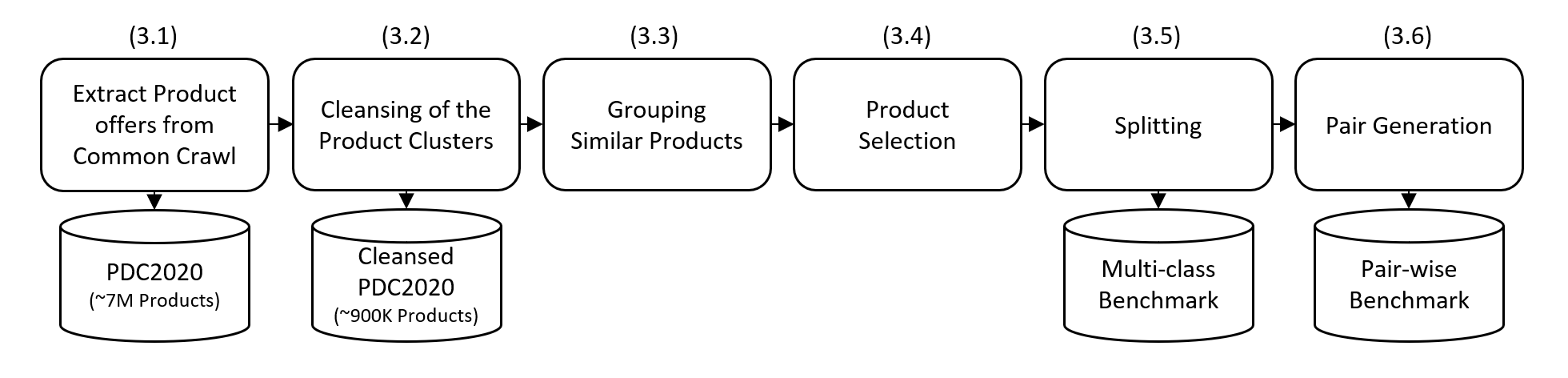}
  \caption {Creation process of WDC Products from the extraction of offers from the Common Crawl to the final benchmark. Numbering at each step refers to the corresponding section in the paper.}
  \label{fig:process}
\end{figure*}

\textbf{Amount of Corner-Cases:} Hard positives and hard negatives are examples that are close to the decision boundary between two or more classes~\cite{simo-serraDiscriminativeLearningDeep2015,schroffFaceNetUnifiedEmbedding2015a,zhanOptimizingDenseRetrieval2021}.
 In the context of WDC Products, we define corner-cases as pairs of matching and non-matching product offers that are dissimilar (matches) or similar (non-matches) with regard to their textual representation. The notion of similarity depends on the chosen similarity metric, which may range from simple string similarity metrics like Jaccard to embedding-based metrics using nearest-neighbor search in the embedding space. Corner-case examples selected in this manner exhibit the property of resembling an example of the respective other class(es). In order to prevent selection bias towards a specific similarity metric, we randomly choose from a set of similarity metrics when determining corner cases (see Section \ref{subsec:building}). 
In the context of product matching, a \textit{positive corner-case} refers to a pair of matching product offers that exhibit dissimilarities in their surface forms, which are usually the result of heterogeneity introduced by different vendors, e.g. mentioning different product features in the offers or using different abbreviations or units of measurement. A \textit{negative corner-case} consists of two non-matching product offers whose textual representations are highly similar, e.g. differ only in a single product feature.

\textbf{Generalization to Unseen Entities:} An unseen entity is an entity that is contained in the test set but not in the training set. Unseen entities are also called out-of-distribution entities~\cite{yangGeneralizedOutofDistributionDetection2022}. In the context of pair-wise entity matching, a test pair may contain descriptions of zero, a single, or two unseen entities, e.g. entities for which no description exists in any training pair. The dimension generalization to unseen entities measures the robustness~\cite{dziugaiteSearchRobustMeasures2020,musaviGeneralizationAbilityNeural1994} of  matching methods to the amount of unseen entities in test pairs. We quantify the robustness of a matcher by comparing precision, recall, and F1 results for test sets containing different amounts of unseen entities. Transfer learning for entity matching~\cite{trabelsiDAMEDomainAdaptation2022,tuDomainAdaptationDeep2022,akbarianrastaghiProbingRobustnessPretrained2022} is closely related to the generalization dimension of the WDC Products benchmark, as in a transfer learning setting most entities in the test set are unseen.

\textbf{Development Set Size:} The  development set consists of all examples (pairs of product offers) that are part of the training set or the validation set. By offering multiple development sets of different sizes, the WDC Product benchmark provides for evaluating the training data efficiency of matching methods.

\textbf{Pairwise- vs. Multi-class Formulation:} The entity matching task has historically been formulated as a binary classification task, meaning that a pair of entity descriptions is classified either as a match or a non-match~\cite{christophides_end--end_2020,BarlaugNeural2021,Christen2012DataMC,elmagarmidDuplicateRecordDetection2007}. For use cases that require matching only a previously known set of entities, it is also possible to approach entity matching as a multi-class classification task. In this case, the classifier labels each entity description with a single label from the label space of all relevant products.

\section{Benchmark Creation}
\label{sec:creation}

The WDC Products benchmark was created using a pipeline consisting of the following six steps: (i) extraction of product offers from the Common Crawl, (ii) cleansing of the product clusters, (iii) grouping similar products, (iv) product selection, (v) splitting, and (vi) pair-generation. This section gives an overview of the six steps which are visualized in Figure \ref{fig:process}. Detailed explanations of every step are given in the following sections.

The first step of the pipeline is the extraction of large amounts of product offers from the Common Crawl\footnote{\url{https://commoncrawl.org/}} using schema.org annotations. Some product offers contain product identifiers like Manufacturer Part Numbers (MPN), Global Trade Item Numbers (GTIN), and Stock-Keeping Units (SKU) which allow us to group offers into clusters containing offers for the same product.  In the second step, we apply several cleansing heuristics to prepare the data for the benchmark. 
In the third step, we group similar product clusters together in order to facilitate the later discovery of corner-cases. Additionally, this grouping allows us to remove some unwanted groups like adult products. In the fourth step, we select several sets of 500 product clusters from the previous grouping to mix and merge for materializing the two dimensions \textit{amount of corner-cases} and \textit{unseen products}. For this purpose, we iterate over the groups and randomly select one seed product cluster per group. To ensure the inclusion of corner cases, we perform similarity searches using different similarity metrics among the product clusters from the previous grouping and the respective selected seed product. For example, to create a set of 500 products with an 80\% corner-case ratio, for each seed product we select 4 highly similar additional product clusters from that group until we have selected 400 products. The remaining 100 are then selected randomly among all groups. For each corner-case selection, we randomly draw from a set of similarity metrics to reduce selection bias. As the basis for the later generation of the unseen dimension, we further select a second distinct set of 500 products for each corner-case ratio.

In the fifth step, the offers in the selected clusters are split into training, validation, and test sets. In order to prevent information leakage in the training process, each offer is only assigned to a single set. Depending on the corner-case ratio, this splitting is either done randomly or using the previously applied similarity metrics to ensure positive corner-cases. At this point, the unseen dimension is materialized as well by replacing e.g. 50\% of the products in the test split with products from the second set of 500 \textit{unseen} products. Finally, the development set size dimension is realized by further selecting subsets from the training split of each product in order to build a medium and small training set. In the sixth and final step, the generated splits are used to generate pairs of offers for the pair-wise formulation of the benchmark. To ensure an appropriate amount of corner-case pairs, we again randomly chose from the same set of similarity metrics that we already used in step 4.

\subsection{Extraction of Product Offers from the Common Crawl}
\label{subsec:extractionCC}

Many e-shops mark up their product offers in HTML pages using schema.org annotations in order to support search engines in extracting the offers and presenting them to their users in the context of e-commerce applications such as Google Shopping\footnote{\url{https://shopping.google.com/}}. In recent years, many of these e-shops have also started to annotate product identifiers within their pages, such as Manufacturer Part Numbers (MPNs), Global Trade
Item Numbers (GTINs), or Stock-Keeping Units (SKUs). These identifiers allow offers for the same product from multiple e-shops to be grouped into clusters after applying some cleansing~\cite{primpeli2019wdc}. The product offers are annotated with attributes such as \textit{name}, \textit{description}, \textit{brand}, and \textit{price}. These annotations offer a unique source of training and evaluation data for entity matching systems. 

The Web Data Commons\footnote{\url{http://webdatacommons.org/structureddata/}} project regularly extracts schema.org annotations from the Common Crawl, the largest web corpus available to the public, in order to monitor the adoption of semantic annotations on the Web and to provide the extracted data for public download.
The WDC Products benchmark uses product offers from the WDC Product Data Corpus V2020 (PDC2020)\footnote{\url{http://webdatacommons.org/largescaleproductcorpus/v2020/}}. The corpus was created by extracting schema.org product data from the September 2020 version of the Common Crawl. The extracted data goes through a pipeline of cleansing steps such as removing offers from listing pages as well as advertisements that are contained in a page in addition to the main offer~\cite{primpeli2019wdc}.  The resulting PDC2020 corpus consists of $\sim$98 million product offers originating from 603,000 websites. These offers all contain some form of product identifier which allows to group them into $\sim$7.1 million clusters having a size of at least two offers. An evaluation of the cleanliness (a cluster is considered clean if all contained offers are for the same real-world product) of the clusters estimated a noise ratio of 6.9\% for clusters having a size smaller or equal to 80 and around 1.8\% for clusters with a size larger than 80. Each product offer in the corpus is described by the textual attributes \textit{title}, \textit{description}, and \textit{brand}, as well as the numerical attribute \textit{price} and a corresponding \textit{priceCurrency} attribute. Its size and the relatively low noise ratio inside product clusters make the PDC2020 corpus a good starting point for the generation of the WDC Products benchmark. 

\subsection{Cleansing of the Product Clusters}
\label{subsec:corpuscleansing}

In order to increase the quality of the product clusters and find a suitable subset for building the benchmark, we apply some further cleansing steps to the PDC2020 corpus. These cleansing steps are detailed below.

\textbf{Language Identification:} PDC2020 is a multi-lingual corpus with product offers originating from all across the Web and thus also includes a lot of non-English product offers. As we want WDC Products to be an English benchmark, we apply some steps to filter out non-English offers. In its original form, PDC2020 contains more than 98M offers and over 7M product clusters with a size of at least two. In the first step, we apply the fastText language identification model\footnote{\url{https://fasttext.cc/docs/en/language-identification.html}} to each row in PDC2020, more specifically to the concatenation of the attributes \textit{title} and \textit{description}. If a product offer has a longer textual description this should be reliably identifiable as English or non-English for the algorithm. For any offers without description, the language identification has to rely on non-English words in the product titles alone. After applying the fastText model, we keep all rows where the classifier confidence is highest for the English language. In the second step, we apply a regular expression to identify non-Latin characters in the remaining product offers and keep only those containing less than four non-Latin characters. We choose to keep these as sometimes product descriptions contain non-Latin characters as part of a model name or branding of the manufacturer. Any product offers with more than four non-Latin characters we assume to be non-English and remove them. 

\textbf{Deduplication and Removal of Short Offers:} In the next step, we concatenate the attributes \textit{title}, \textit{description}, and \textit{brand} and drop any duplicate rows on this combined attribute, keeping only the first occurrence.  Finally, we remove all product offers where the title attribute contains less than five tokens, as we expect offers with such short titles to be unsuited due to being very sparsely described which results in unsolvable ambiguity when trying to match similarly sparse but different products.

\textbf{Outlier Removal:} In the final step, we further aim to eliminate any remaining noisy products in the product clusters. For this purpose, we employ a heuristic based on simple word occurrence among offers inside a cluster. More specifically, we scan each clusters offers and keep track of general \textit{title} length while building a dictionary of word counts across offers' \textit{titles}. We expect any offer containing very unique words compared to all others in the cluster to be noisy non-matching product offers and subsequently remove them from the cluster and corpus.

After the cleansing steps, PDC2020 contains 22M offers and 900K product clusters of size greater than one. This cleansed corpus is used as the base for the creation of WDC Products and is also available for download on the WDC Products website together with the creation code. In combination, these two artifacts allow an interested user to create more fine-grained steps for each of the benchmark dimensions as well as the generation of entirely new benchmarks similar to WDC Products.

\subsection{Grouping Similar Products}
\label{subsec:clustering}

To group similar products for easier discovery of corner-case products in later stages, we first employ the scikit-learn implementation of DBSCAN clustering\footnote{\url{https://scikit-learn.org/stable/modules/generated/sklearn.cluster.DBSCAN.html}} on the data corpus with an epsilon of 0.35 and min\_samples of 1 to determine coarse groups of similar products which we can later use for corner-case selection. As feature vector for each product, we use simple binary word occurrence after lower-casing and removing tags and punctuation. The values for epsilon and min\_samples were chosen as to generate the largest amount of groups containing products with at least 7 offers. The minimum number of 7 is chosen as we need at least this amount for cleanly splitting offers into train validation and test as described in section \ref{subsec:building}. A random manual check of the generated groups showed that they contain either only highly similar products, e.g. hard drives from the same vendor, or very similar products, e.g. graphics cards from the same but also other vendors. In the next step, we split the corpus with annotated DBSCAN groups into two parts, the first one (from which we draw the \textit{seen} portion of the benchmark, represented by 629 DBSCAN groups) containing all products represented by at least 7 offers and the second one (from which we draw the \textit{unseen} examples, represented by 2,845 DBSCAN groups) with products having between 2 and 6 offers. In the final step, two domain experts manually check all 629 groups in the first part and annotate them with \textit{useful} or \textit{avoid} depending on the cleanliness of the group and the suitability of the products therein for the benchmark, e.g. we make the decision to exclude adult products. The domain experts then look at the same number of groups for the second part and annotate them in an analogous manner. The results of this process are two sets of grouped and manually curated products from which we select sets of products in the next step.

\begin{figure}
  \centering
  \includegraphics[width=\columnwidth]{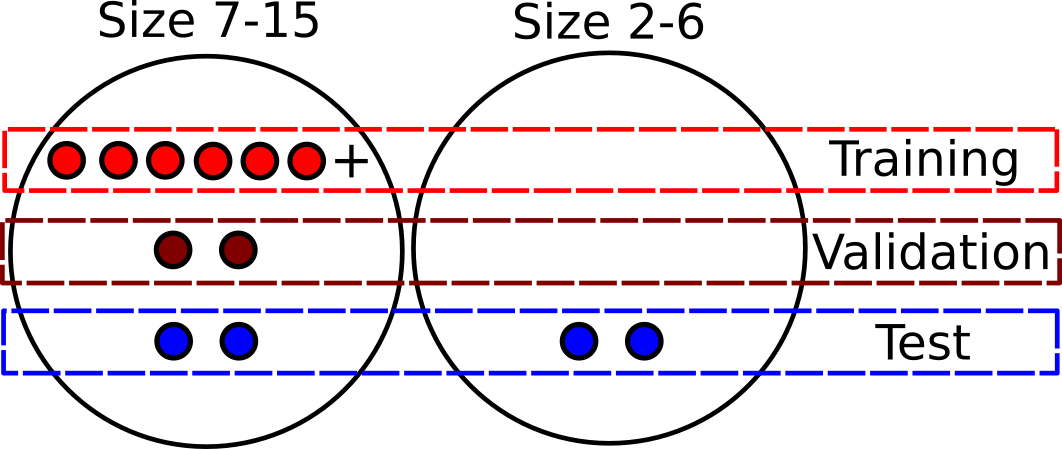}
  \caption{Depiction of the cluster sizes and distribution of offers into splits for seen product clusters with size 7-15 (left) and unseen product clusters with size 2-6 (right)}
  \label{fig:clustersizes}
\end{figure}

\subsection{Product Selection}
\label{subsec:building}

For the creation of WDC Products, we select multiple sets of 500 products from the cleansed and grouped PDC2020 corpus along three dimensions that characterize the final benchmark. The three dimensions are: (i) amount of corner-cases, (ii) amount of unseen products, and (iii) development set size. The amount of corner-cases determines the percentage of the 500 products which have at least 4 textually highly similar products in the set of 500. Thus, the higher the amount of corner-cases in a dataset, the harder it will be for matching systems to disambiguate between matching and non-matching offers as textual descriptions will be very similar for such products. We vary this dimension in 3 steps, 80\%, 50\%, and 20\%. The amount of unseen products is only relevant for test splits and determines the percentage of products that are represented in the training and validation sets. We vary this dimension from 0\% unseen over 50\% to 100\% unseen. The third dimension varies the amount of available development data in sizes small, medium, and large. These sets of products need to fulfill two characteristics: each product must be represented by at least 7 (seen) or 2 (unseen) unique product offers for the purpose of cleanly splitting them among training, validation, and test sets and at least four very similar but different products need to be available in the selected set of products for the purpose of generating negative corner-cases.

In the following, we describe the selection of 500 products for the final benchmark datasets but keep the corner-case dimension fixed to 80\% corner-cases to improve understandability. As the process is analogous to the other corner-case percentages, the following description can be directly applied to them as well.

We select 400 products which include at least 4 negative corner-cases for each product (80\% corner-cases) and 100 random products (the remaining 20\%) once from the seen and once from the unseen groups of our corpus so we can later build sets for each unseen percentage. We begin by iterating through the annotated DBSCAN groups. Inside each DBSCAN group, we randomly select one product cluster and then search for the four most similar product clusters inside that group by randomly alternating between the most similar examples on the product \textit{title} according to a variety of similarity metrics: \textit{Cosine}, \textit{DICE} and \textit{Generalized Jaccard} similarities from the py\_stringmatching\footnote{\url{https://github.com/anhaidgroup/py\_stringmatching}} package and a fastText embedding model trained on the titles of the product matching benchmarks \textit{Amazon-Google}, \textit{Abt-Buy}, and \textit{Walmart-Amazon}. While we could use a stricter version of the DBSCAN grouping to select corner-cases, this would limit the selection to the representation and similarity metric used in this clustering which would essentially lead to a benchmark that can be easily solved using the DBSCAN algorithm. By using a varied selection of similarity metrics, we avoid biasing the benchmark towards a single method of corner-case selection while ensuring it cannot be solved by a matcher based on a single metric. After collecting 400 products in this way once for the \textit{seen} and once for the \textit{unseen} split, we use the remaining products and randomly select a further 100 for seen and unseen, bringing both product sets to 500 products with a 80\% negative corner-case ratio.

\begin{table*}[htb]
\centering
\caption{Statistics of the training, validation and test sets of WDC Products along the dimensions amount of corner-cases and development set size for the pair-wise and multi-class matching tasks.}
\label{tab:sizestats}
\resizebox{2\columnwidth}{!}{%
\begin{tabular}{@{}l|c|ccccccccc|ccc@{}}
\toprule
Type       & Corner-Cases          & \multicolumn{9}{c|}{Pair-Wise}                                                                                     & \multicolumn{3}{c}{Multi-Class} \\ \midrule
           &                       & \multicolumn{3}{c|}{Small}               & \multicolumn{3}{c|}{Medium}                & \multicolumn{3}{c|}{Large} & Small    & Medium    & Large    \\
           &                       & All   & Pos & \multicolumn{1}{c|}{Neg}   & All   & Pos   & \multicolumn{1}{c|}{Neg}   & All     & Pos    & Neg     &          &           &          \\ \midrule
Training   & \multirow{3}{*}{80\%} & 2,500 & 500 & \multicolumn{1}{c|}{2,000} & 6,000 & 1,500 & \multicolumn{1}{c|}{4,500} & 19,835  & 8,471  & 11,364  & 1,000    & 1,500     & 2,841    \\
Validation &                       & 2,500 & 500 & \multicolumn{1}{c|}{2,000} & 3,500 & 500   & \multicolumn{1}{c|}{3,000} & 4,500   & 500    & 4,000   & 1,000    & 1,000     & 1,000    \\
Test       &                       & 4,500 & 500 & \multicolumn{1}{c|}{4,000} & 4,500 & 500   & \multicolumn{1}{c|}{4,000} & 4,500   & 500    & 4,000   & 1,000    & 1,000     & 1,000    \\ \midrule
Training   & \multirow{3}{*}{50\%} & 2,500 & 500 & \multicolumn{1}{c|}{2,000} & 6,000 & 1,500 & \multicolumn{1}{c|}{4,500} & 19,607  & 8,339  & 11,268  & 1,000    & 1,500     & 2,817    \\
Validation &                       & 2,500 & 500 & \multicolumn{1}{c|}{2,000} & 3,500 & 500   & \multicolumn{1}{c|}{3,000} & 4,500   & 500    & 4,000   & 1,000    & 1,000     & 1,000    \\
Test       &                       & 4,500 & 500 & \multicolumn{1}{c|}{4,000} & 4,500 & 500   & \multicolumn{1}{c|}{4,000} & 4,500   & 500    & 4,000   & 1,000    & 1,000     & 1,000    \\ \midrule
Training   & \multirow{3}{*}{20\%} & 2,500 & 500 & \multicolumn{1}{c|}{2,000} & 6,000 & 1,500 & \multicolumn{1}{c|}{4,500} & 19,015  & 7,963  & 11,052  & 1,000    & 1,500     & 2,763    \\
Validation &                       & 2,500 & 500 & \multicolumn{1}{c|}{2,000} & 3,500 & 500   & \multicolumn{1}{c|}{3,000} & 4,500   & 500    & 4,000   & 1,000    & 1,000     & 1,000    \\
Test       &                       & 4,500 & 500 & \multicolumn{1}{c|}{4,000} & 4,500 & 500   & \multicolumn{1}{c|}{4,000} & 4,500   & 500    & 4,000   & 1,000    & 1,000     & 1,000    \\ \bottomrule
\end{tabular}%
}
\end{table*}

\subsection{Splitting}
\label{subsec:splitting}

In this step, we split the offers of each of the 500 product clusters of the \textit{seen} split to build training, validation, and test offers. For the seen products, we limit the maximum amount of offers we select from a product cluster to 15. For the unseen products, we sample exactly 2 offers from the respective cluster. Figure \ref{fig:clustersizes} shows how we proceed with splitting the selected offers into training validation and test offers. 
Validation and Test splits are assigned two offers each, the remaining offers are part of the training split. The selection of offers for each split is done along the corner-case dimension, meaning for 80\% of products we use the same similarity metrics as before for all combinations of offers in a product cluster and sort them by increasing similarity to find positive corner-cases. We then slice this list at the first fifth and randomly select two pairs from the corner-case side (left) for inclusion in test and validation splits. The rest are assigned to the training split.

To build the unseen dimension we proceed with systematically replacing products in the fully seen test set with products from the unseen split. Care is taken to keep the corner-case amount the same, i.e. for building the 50\% unseen test set, we need to replace 250 products with unseen products in a ratio of 80\% corner-cases and 20\% random. Finally, to build the 100\% unseen test set, we can simply fully replace the original test set with the corresponding selection from the unseen split. In summary, we generate three test sets (0\%, 50\%, and 100\% unseen) for the corner-case ratio of 80\%.

Finally, we generate datasets for the last dimension, the development set size. For this purpose we split the subsets of each product that were assigned to the training split as follows: All product offers are assigned to the large training set, three among those are assigned to the medium training set and two of the three are finally assigned to the small training set. For 80\% of products, we apply the same positive corner-case procedure as before to ensure pairs in the small and medium training sets are corner-cases.

As stated before, we proceed with the same process for the corner-case ratios 50\% and 20\%. Finally, we have produced three test sets and three development sets for each corner-case ratio - nine test sets, nine validation sets, and nine training sets overall. We can directly use these datasets for benchmarking multi-class entity matchers. For pair-wise matching, we still need to generate pairs from the selected product offers.

\subsection{Pair Generation}
\label{subsec:pairgen}

To generate the pair-wise datasets from the selected products and splits we can directly use the previously prepared training, validation, and test splits. We iterate over each offer in each set and build all positive pairs as well as varying amounts of negative corner-cases, and random pairs depending on the development set size. In the following, we use the building process of the development size \textit{large} as an example.

First, we iterate over each product cluster in the selected training set. A product cluster in the large training set can contain between 3 and 11 unique offers. We generate all possible positive pairs using the available offers and include them in the final large training set. Afterward, we iterate over each offer in the product cluster and select for each the 3 most similar (using alternating similarities as before) product offers among the remaining 499 product clusters in the dataset. If a pair is already contained in the final training set, e.g. because of a mirrored pair, we select the next most similar pair to add instead. Finally, we add one random negative pair for a total of 4 negative pairs for each offer in the current product cluster. We continue this process for all 499 remaining product clusters resulting in the final pair-wise large training set.

The generation of all pair-wise test sets and the large validation sets follow the same process as for the large training sets. For the medium and small training and validation sets, we reduce the amount of selected negative corner-cases per product offer to 2 (medium sets) and 1 (small sets) to simulate a reduced labeling effort in addition to the lower amount of available product offers per product for the different training sets (3 for medium and 2 for small). 

\begin{table*}[htb]
\centering
\caption{Attribute density and length statistics across the merged sets by development set size and amount of corner-cases.}
\label{tab:attributes}
\resizebox{2\columnwidth}{!}{%
\begin{tabular}{@{}l|c|c|ccccc|cc@{}}
\toprule
\begin{tabular}[c]{@{}l@{}}Development\\ Set Size\end{tabular} & Corner-Cases          & \# Entities & \multicolumn{5}{c|}{\begin{tabular}[c]{@{}c@{}}Density (\%)/ \\ Median Length (\# Words)\end{tabular}} & \multicolumn{2}{c}{\begin{tabular}[c]{@{}c@{}}Vocabulary\\ (unique)\end{tabular}} \\ \midrule
                                                               & \multicolumn{1}{l|}{} &             & title            & description           & price           & priceCurrency           & brand           & Words                                   & Tokens                                 \\ \midrule
Average                                                        & -                     & -           & 100/8            & 75/32                 & 93/1            & 90/1                    & 35/1            & 18,832                                  & 11,876                                  \\ \midrule
Small                                                          & \multirow{3}{*}{80\%} & 500         & 100/8            & 75/37                 & 94/1            & 91/1                    & 34/1            & 16,662                                  & 11,420                                  \\
Medium                                                         &                       & 500         & 100/8            & 75/37                 & 94/1            & 91/1                    & 34/1            & 17,612                                  & 11,718                                  \\
Large                                                          &                       & 500         & 100/8            & 75/36                 & 94/1            & 91/1                    & 34/1            & 19,862                                  & 12,246                                  \\ \midrule
Small                                                          & \multirow{3}{*}{50\%} & 500         & 100/8            & 76/32                 & 93/1            & 90/1                    & 35/1            & 17,232                                  & 11,548                                  \\
Medium                                                         &                       & 500         & 100/8            & 76/32                 & 93/1            & 90/1                    & 35/1            & 18,313                                  & 11,912                                  \\
Large                                                          &                       & 500         & 100/8            & 75/32                 & 93/1            & 90/1                    & 35/1            & 20,361                                  & 12,372                                  \\ \midrule
Small                                                          & \multirow{3}{*}{20\%} & 500         & 100/8            & 74/29                 & 92/1            & 90/1                    & 35/1            & 16,179                                  & 11,201                                  \\
Medium                                                         &                       & 500         & 100/8            & 74/29                 & 92/1            & 90/1                    & 35/1            & 17,220                                  & 11,563                                  \\
Large                                                          &                       & 500         & 100/8            & 74/29                 & 92/1            & 90/1                    & 34/1            & 19,235                                  & 12,028                                  \\ \bottomrule
\end{tabular}%
}
\end{table*}

\section{Benchmark Profiling}
\label{sec:profiling}

 \textbf{General Statistics and Size:} The WDC Products benchmark consists of overall 11,715 unique product offers which describe 2,162 different products. It further consists of overall nine training, nine validation, and nine test sets each for pair-wise matching as well as multi-class matching. Table \ref{tab:sizestats} shows statistics about the size of the datasets for each of these splits. Each split contains offers for exactly 500 products. All test sets contain exactly 4,500 pairs of offers (pair-wise) or 1,000 offers (multi-class). The sizes of the training and validation sets on the other hand vary across the development set size dimension from 5,000 (pairwise) and 2,000 (multi-class), representing the small dataset, to 8500/2,500 in the medium and $\sim$25,000/$\sim$4,000 in the large set. This allows for the evaluation of matching systems along the dimension of development set size while ensuring comparability between pair-wise and multi-class tasks, as both datasets always contain the same set of offers in training, validation, and test as well as no overlapping offers between the training and evaluation splits. In addition to the development set size dimension, each dataset exists in three versions of increasing difficulty signified by the amount of corner-cases in the contained products. Finally, for each of the three difficulties, three test sets of the same size are available with an increasing amount of unseen products from 0\% over 50\% to 100\% representing the third and final dimension of the benchmark. As explained in Section \ref{sec:creation} when replacing seen with unseen products, care is taken to preserve the corresponding corner-case ratio.

\textbf{Attribute Density and Vocabulary:} Each offer in WDC Products has five attributes, \textit{title}, \textit{description}, \textit{price}, \textit{priceCurrency}, and \textit{brand}. Table \ref{tab:attributes} contains statistics about density and value length for each attribute across all datasets. Most attributes have a density of over 90\% with \textit{description} (75\%) only filled in three-quarters of offers and \textit{brand} (35\%) being the least dense attribute. Though manual inspection revealed that brand information is often found in the \textit{title} attribute if the \textit{brand} attribute itself is missing. The median value lengths of the attributes reveal that the \textit{title} can be considered a shorter textual attribute while the \textit{description} contains longer strings. \textit{Brand} and \textit{priceCurrency} mainly consist of a single word which in the case of \textit{priceCurrency} is often a three-character identifier for the currency, such as \textit{USD} or \textit{EUR}. The \textit{price} represents the single mainly numerical attribute of the benchmark. An example of each attribute can be seen in Figure \ref{fig:example}. Table \ref{tab:attributes} further shows statistics for the unique vocabulary used across all datasets highlighting the heterogeneity of the benchmark with each dataset containing between 17K and 20K unique words on average as well as making use of around 12K (24\%) tokens contained in RoBERTa's vocabulary of size $\sim$50K.

\textbf{Label Quality:} As the labels of the benchmark originate from the automatic clustering of offers using annotated product identifiers from the web, we perform a manual evaluation of their correctness on the test splits of WDC Products. Two expert annotators check the match and non-match labels from a sample of labeled pairs which are sampled from all nine available test splits. From each test split we sample an equal amount of positives and negatives while the overall amount depends on the corner-case percentage of the respective test set. As the test sets with higher corner-case ratios contain harder-to-match pairs this ensures the sampling does not favor easily disambiguated pairs. Specifically,  we sample 100/60/40 pairs for corner-case percentages of 80/50/20 resulting in 600 (300 positives, 300 negatives) labeled pairs in total. Overall, the noise level in the sample is estimated as 4.00\% by annotator one and 4.17\% by annotator two with a Cohen's Kappa of 0.91.

\begin{table*}[htb]
\centering
\caption{Results of the pair-wise experiments over all three dimensions, amount of corner-cases, amount of unseen products in the test set, and development set size. Results are F1 scores for the class match. Bold results indicate the best F1 score for that benchmark variant, underlined indicates second best.}
\label{tab:resultspair}
\resizebox{2.1\columnwidth}{!}{%
\begin{tabular}{@{}l|c|ccc|ccc|ccc|ccc|ccc|ccc@{}}
\toprule
\begin{tabular}[c]{@{}l@{}}Development\\ Set Size\end{tabular} &
  Corner Cases &
  \multicolumn{3}{c|}{Word-Cooc} &
  \multicolumn{3}{c|}{Magellan} &
  \multicolumn{3}{c|}{RoBERTa} &
  \multicolumn{3}{c|}{Ditto} &
  \multicolumn{3}{c|}{HierGAT} &
  \multicolumn{3}{c}{R-SupCon} \\ \midrule
 &
  \multicolumn{1}{l|}{} &
  Seen &
  Half-Seen &
  Unseen &
  Seen &
  Half-Seen &
  Unseen &
  Seen &
  Half-Seen &
  Unseen &
  Seen &
  Half-Seen &
  Unseen &
  Seen &
  Half-Seen &
  Unseen &
  Seen &
  Half-Seen &
  Unseen \\ \midrule
Small &
  \multirow{3}{*}{80\%} &
  43.73 &
  40.07 &
  27.46 &
  31.15 &
  33.75 &
  33.34 &
  {\ul 65.45} &
  \textbf{66.68} &
  \textbf{64.50} &
  58.33 &
  58.97 &
  57.16 &
  59.65 &
  61.54 &
  {\ul 60.63} &
  \textbf{77.48} &
  {\ul 64.25} &
  51.91 \\
Medium &
   &
  52.66 &
  44.06 &
  30.57 &
  30.55 &
  35.00 &
  33.47 &
  72.18 &
  {\ul 72.05} &
  \textbf{70.13} &
  {\ul 74.07} &
  \textbf{72.78} &
  {\ul 69.49} &
  71.40 &
  67.64 &
  67.45 &
  \textbf{79.99} &
  67.21 &
  53.10 \\
Large &
   &
  56.67 &
  50.24 &
  30.26 &
  31.96 &
  36.42 &
  34.95 &
  78.15 &
  \textbf{75.52} &
  \textbf{69.75} &
  {\ul 79.46} &
  68.81 &
  67.94 &
  75.42 &
  {\ul 73.20} &
  {\ul 68.53} &
  \textbf{82.15} &
  67.27 &
  53.31 \\ \midrule
Small &
  \multirow{3}{*}{50\%} &
  48.10 &
  40.23 &
  29.44 &
  31.38 &
  32.44 &
  33.34 &
  68.69 &
  \textbf{69.18} &
  \textbf{65.79} &
  {\ul 70.19} &
  65.40 &
  {\ul 61.84} &
  61.70 &
  60.74 &
  59.21 &
  \textbf{78.43} &
  {\ul 68.24} &
  57.44 \\
Medium &
   &
  58.07 &
  46.04 &
  29.70 &
  35.83 &
  37.45 &
  36.61 &
  78.58 &
  \textbf{75.91} &
  \textbf{71.14} &
  {\ul 79.16} &
  {\ul 75.22} &
  {\ul 70.24} &
  75.17 &
  73.30 &
  68.74 &
  \textbf{81.88} &
  68.69 &
  57.23 \\
Large &
   &
  60.39 &
  51.15 &
  31.64 &
  35.41 &
  37.39 &
  38.51 &
  82.46 &
  {\ul 78.89} &
  \textbf{71.52} &
  {\ul 83.88} &
  \textbf{79.36} &
  69.36 &
  81.47 &
  76.98 &
  {\ul 71.34} &
  \textbf{85.16} &
  71.15 &
  57.68 \\ \midrule
Small &
  \multirow{3}{*}{20\%} &
  46.55 &
  45.30 &
  33.30 &
  34.17 &
  37.50 &
  35.18 &
  {\ul 75.24} &
  \textbf{75.87} &
  {\ul 72.44} &
  73.96 &
  {\ul 75.36} &
  \textbf{72.62} &
  64.34 &
  64.62 &
  68.25 &
  \textbf{85.06} &
  73.09 &
  64.56 \\
Medium &
   &
  58.04 &
  51.33 &
  34.38 &
  36.90 &
  40.68 &
  37.10 &
  {\ul 83.68} &
  \textbf{80.60} &
  \textbf{78.35} &
  83.43 &
  {\ul 78.40} &
  {\ul 76.33} &
  79.53 &
  77.60 &
  74.84 &
  \textbf{87.46} &
  73.17 &
  63.52 \\
Large &
   &
  61.81 &
  54.26 &
  35.83 &
  37.58 &
  41.57 &
  37.23 &
  {\ul 87.80} &
  {\ul 82.17} &
  \textbf{78.64} &
  87.52 &
  \textbf{82.81} &
  {\ul 77.92} &
  84.15 &
  79.54 &
  75.53 &
  \textbf{89.04} &
  74.59 &
  62.45 \\ \bottomrule
\end{tabular}%
}
\end{table*}
\begin{table*}[htb]
\centering
\caption{Precision and recall values for the neural matching systems.}
\label{tab:resultspair_pr}
\resizebox{2.1\columnwidth}{!}{%
\begin{tabular}{@{}l|c|cccccc|cccccc|cccccc|cccccc@{}}
\toprule
\begin{tabular}[c]{@{}l@{}}Development\\ Set Size\end{tabular} &
  Corner Cases &
  \multicolumn{6}{c|}{RoBERTa} &
  \multicolumn{6}{c|}{Ditto} &
  \multicolumn{6}{c|}{HierGAT} &
  \multicolumn{6}{c}{R-SupCon} \\ \midrule
 &
  \multicolumn{1}{l|}{} &
  \multicolumn{2}{c|}{Seen} &
  \multicolumn{2}{c|}{Half-Seen} &
  \multicolumn{2}{c|}{Unseen} &
  \multicolumn{2}{c|}{Seen} &
  \multicolumn{2}{c|}{Half-Seen} &
  \multicolumn{2}{c|}{Unseen} &
  \multicolumn{2}{c|}{Seen} &
  \multicolumn{2}{c|}{Half-Seen} &
  \multicolumn{2}{c|}{Unseen} &
  \multicolumn{2}{c|}{Seen} &
  \multicolumn{2}{c|}{Half-Seen} &
  \multicolumn{2}{c}{Unseen} \\ \midrule
 &
  \multicolumn{1}{l|}{} &
  P &
  \multicolumn{1}{c|}{R} &
  P &
  \multicolumn{1}{c|}{R} &
  P &
  R &
  P &
  \multicolumn{1}{c|}{R} &
  P &
  \multicolumn{1}{c|}{R} &
  P &
  R &
  P &
  \multicolumn{1}{c|}{R} &
  P &
  \multicolumn{1}{c|}{R} &
  P &
  R &
  P &
  \multicolumn{1}{c|}{R} &
  P &
  \multicolumn{1}{c|}{R} &
  P &
  R \\ \midrule
Small &
  \multirow{3}{*}{80\%} &
  55.84 &
  \multicolumn{1}{c|}{79.07} &
  57.06 &
  \multicolumn{1}{c|}{80.27} &
  54.20 &
  79.80 &
  46.28 &
  \multicolumn{1}{c|}{79.33} &
  46.28 &
  \multicolumn{1}{c|}{83.13} &
  43.93 &
  83.60 &
  49.20 &
  \multicolumn{1}{c|}{76.27} &
  51.06 &
  \multicolumn{1}{c|}{77.53} &
  48.92 &
  80.13 &
  71.19 &
  \multicolumn{1}{c|}{85.00} &
  55.87 &
  \multicolumn{1}{c|}{75.60} &
  40.14 &
  73.47 \\
Medium &
   &
  66.61 &
  \multicolumn{1}{c|}{79.07} &
  65.60 &
  \multicolumn{1}{c|}{80.47} &
  61.92 &
  81.40 &
  70.97 &
  \multicolumn{1}{c|}{77.53} &
  67.10 &
  \multicolumn{1}{c|}{79.53} &
  60.76 &
  81.47 &
  68.68 &
  \multicolumn{1}{c|}{74.40} &
  61.08 &
  \multicolumn{1}{c|}{75.93} &
  59.82 &
  77.47 &
  76.18 &
  \multicolumn{1}{c|}{84.20} &
  60.08 &
  \multicolumn{1}{c|}{76.27} &
  41.73 &
  73.00 \\
Large &
   &
  73.82 &
  \multicolumn{1}{c|}{83.07} &
  69.73 &
  \multicolumn{1}{c|}{82.53} &
  61.93 &
  80.27 &
  77.09 &
  \multicolumn{1}{c|}{81.33} &
  63.81 &
  \multicolumn{1}{c|}{74.80} &
  59.67 &
  78.93 &
  71.42 &
  \multicolumn{1}{c|}{80.00} &
  67.27 &
  \multicolumn{1}{c|}{80.47} &
  60.96 &
  78.60 &
  79.09 &
  \multicolumn{1}{c|}{85.47} &
  59.89 &
  \multicolumn{1}{c|}{76.73} &
  42.32 &
  72.00 \\ \midrule
Small &
  \multirow{3}{*}{50\%} &
  60.35 &
  \multicolumn{1}{c|}{79.93} &
  58.77 &
  \multicolumn{1}{c|}{84.33} &
  56.00 &
  80.00 &
  62.43 &
  \multicolumn{1}{c|}{80.20} &
  54.06 &
  \multicolumn{1}{c|}{83.20} &
  50.15 &
  82.20 &
  51.83 &
  \multicolumn{1}{c|}{76.67} &
  47.39 &
  \multicolumn{1}{c|}{84.60} &
  45.83 &
  83.73 &
  73.56 &
  \multicolumn{1}{c|}{84.00} &
  57.00 &
  \multicolumn{1}{c|}{85.00} &
  45.83 &
  76.93 \\
Medium &
   &
  76.02 &
  \multicolumn{1}{c|}{81.33} &
  67.92 &
  \multicolumn{1}{c|}{86.07} &
  62.61 &
  82.40 &
  75.99 &
  \multicolumn{1}{c|}{82.87} &
  65.28 &
  \multicolumn{1}{c|}{88.87} &
  60.74 &
  83.33 &
  74.72 &
  \multicolumn{1}{c|}{75.80} &
  65.32 &
  \multicolumn{1}{c|}{83.73} &
  59.10 &
  82.40 &
  79.04 &
  \multicolumn{1}{c|}{84.93} &
  57.58 &
  \multicolumn{1}{c|}{85.13} &
  45.70 &
  76.53 \\
Large &
   &
  83.20 &
  \multicolumn{1}{c|}{81.80} &
  71.89 &
  \multicolumn{1}{c|}{87.53} &
  63.26 &
  82.40 &
  84.29 &
  \multicolumn{1}{c|}{83.53} &
  72.21 &
  \multicolumn{1}{c|}{88.13} &
  59.25 &
  83.67 &
  80.49 &
  \multicolumn{1}{c|}{87.47} &
  68.60 &
  \multicolumn{1}{c|}{87.73} &
  62.01 &
  84.00 &
  84.08 &
  \multicolumn{1}{c|}{86.27} &
  59.81 &
  \multicolumn{1}{c|}{87.80} &
  47.07 &
  74.47 \\ \midrule
Small &
  \multirow{3}{*}{20\%} &
  71.72 &
  \multicolumn{1}{c|}{79.47} &
  69.60 &
  \multicolumn{1}{c|}{83.53} &
  63.66 &
  84.13 &
  67.13 &
  \multicolumn{1}{c|}{82.87} &
  68.02 &
  \multicolumn{1}{c|}{84.80} &
  63.55 &
  85.13 &
  56.10 &
  \multicolumn{1}{c|}{76.60} &
  54.95 &
  \multicolumn{1}{c|}{79.33} &
  58.86 &
  81.33 &
  83.09 &
  \multicolumn{1}{c|}{87.13} &
  63.60 &
  \multicolumn{1}{c|}{85.93} &
  53.59 &
  81.20 \\
Medium &
   &
  83.43 &
  \multicolumn{1}{c|}{83.93} &
  75.12 &
  \multicolumn{1}{c|}{87.00} &
  70.63 &
  88.00 &
  83.10 &
  \multicolumn{1}{c|}{83.80} &
  71.70 &
  \multicolumn{1}{c|}{86.53} &
  68.54 &
  86.13 &
  77.26 &
  \multicolumn{1}{c|}{81.93} &
  71.66 &
  \multicolumn{1}{c|}{84.67} &
  66.95 &
  84.87 &
  87.05 &
  \multicolumn{1}{c|}{87.87} &
  63.57 &
  \multicolumn{1}{c|}{86.20} &
  52.96 &
  79.33 \\
Large &
   &
  87.81 &
  \multicolumn{1}{c|}{87.80} &
  75.53 &
  \multicolumn{1}{c|}{90.13} &
  70.51 &
  88.93 &
  89.93 &
  \multicolumn{1}{c|}{85.27} &
  78.59 &
  \multicolumn{1}{c|}{87.53} &
  72.34 &
  84.47 &
  81.17 &
  \multicolumn{1}{c|}{87.40} &
  71.35 &
  \multicolumn{1}{c|}{89.93} &
  66.49 &
  87.47 &
  88.69 &
  \multicolumn{1}{c|}{89.40} &
  64.80 &
  \multicolumn{1}{c|}{87.87} &
  51.13 &
  80.20 \\ \bottomrule
\end{tabular}%
}
\end{table*}
\section{Experimental Evaluation}
\label{sec:experiments}

In order to test the WDC Products benchmark and showcase the usefulness of its multi-dimensional design, we evaluate the following supervised entity matching methods using the benchmark: a word (co-)occurrence baseline, Magellan~\cite{kondaMagellanBuildingEntity2016a}, Ditto~\cite{liDeepEntityMatching2020}, RoBERTa-base~\cite{liu_roberta_2019}, the same RoBERTa model using supervised contrastive pre-training (R-SupCon~\cite{peetersSupervisedContrastiveLearning2022a}), and a matcher based on hierarchical token aggregation (HierGAT~\cite{yaoEntityResolutionHierarchical2022}). All models are trained three times per setup and we report the average F1.

\subsection{Matching Systems}
\label{subsec:baselines}

\textbf{Word (Co-)Occurrence:} The Word (Co-)Occurrence model represents a simple symbolic baseline for both pair-wise and multi-class matching. It uses the binary word co-occurrence between two entity descriptions in a pair as feature input to a binary LinearSVM sklearn~\cite{pedregosaScikitlearnMachineLearning2011} classifier. For the multi-class matching case, the feature input is a binary word occurrence vector instead. We perform a grid search over various parameter combinations during training.

\textbf{Magellan:} We employ Magellan~\cite{kondaMagellanBuildingEntity2016a} as a second symbolic baseline. Magellan selects appropriate similarity metrics for each attribute depending on the attribute datatype. The resulting attribute-specific similarity scores are input to a sklearn binary Random Forest classifier which determines the matching decision. Similarly to the previous baseline, we optimize model parameters using a grid search.

\textbf{RoBERTa:} The RoBERTa-base~\cite{liu_roberta_2019} model is the first sub-symbolic baseline representing a Transformer-based language model which has been shown to reach high performance on the entity matching task~\cite{liDeepEntityMatching2020,peetersDualobjectiveFinetuningBERT2021}.
We fine-tune the RoBERTa model for both pair-wise and multi-class matching for 50 epochs with early stopping after 10 epochs if the validation score does not improve. The batch size is set to 64 and the learning rate is linearly decreasing with warmup with a maximum value of 5e-5.

\textbf{Ditto:} The Ditto~\cite{liDeepEntityMatching2020} matching system is one of the first entity matching systems using Transformer language models. Ditto introduces various data augmentation and domain knowledge injection modules that can be added for model training. For our experiments, we use a RoBERTa-base language model and activate the data augmentation module, specifically the \textit{delete} operator.
We fine-tune the Ditto model for 50 epochs with a batch size of 64 and a learning rate of 5e-5 on a linearly decreasing schedule with warmup.

\textbf{R-SupCon:} The R-SupCon~\cite{peetersSupervisedContrastiveLearning2022a} model is trained in two stages, a supervised contrastive~\cite{khoslaSupervisedContrastiveLearning2021} pre-training stage followed by a cross-entropy-based fine-tuning stage.
After the contrastive pre-training, we freeze the parameters of the RoBERTa encoder and only fine-tune the classification head with cross-entropy loss.
We pre-train the R-SupCon model for 200 epochs using a batch size of 1024 and linearly decreasing learning rate with warmup with a maximum value of 5e-5. The parameters for the fine-tuning stage correspond to the ones of the RoBERTa baseline.

\textbf{HierGAT:} The HierGAT~\cite{yaoEntityResolutionHierarchical2022} model is a recently proposed model combining the Transformer language model's attention mechanism and a hierarchical graph attention network. We train the HierGAT model for 50 epochs. We use a batch size of 16 and a linearly decreasing learning rate of 5e-6 with warmup.

\subsection{Pairwise Benchmark Results}
\label{subsec:results}

\begin{figure}
  \centering
  \includegraphics[width=\columnwidth]{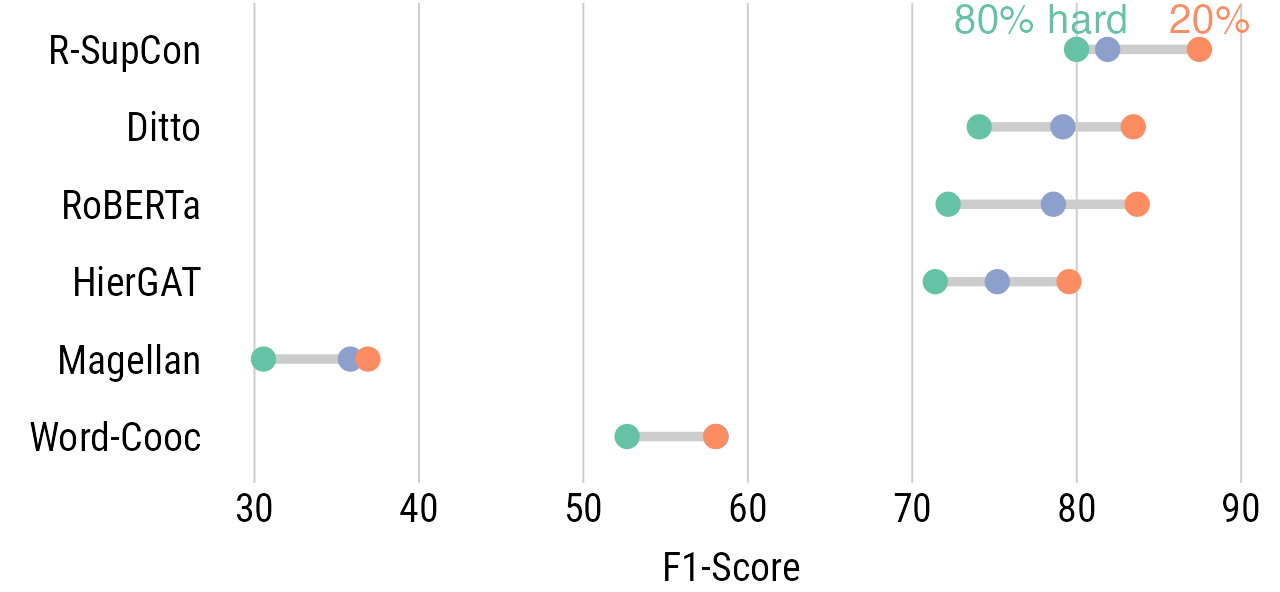}
  \caption {Varying corner-case ratio combined with no unseen entities in the test set and development set size medium.}
  \label{fig:hardness}
\end{figure}

\begin{figure}
  \centering
  \includegraphics[width=\columnwidth]{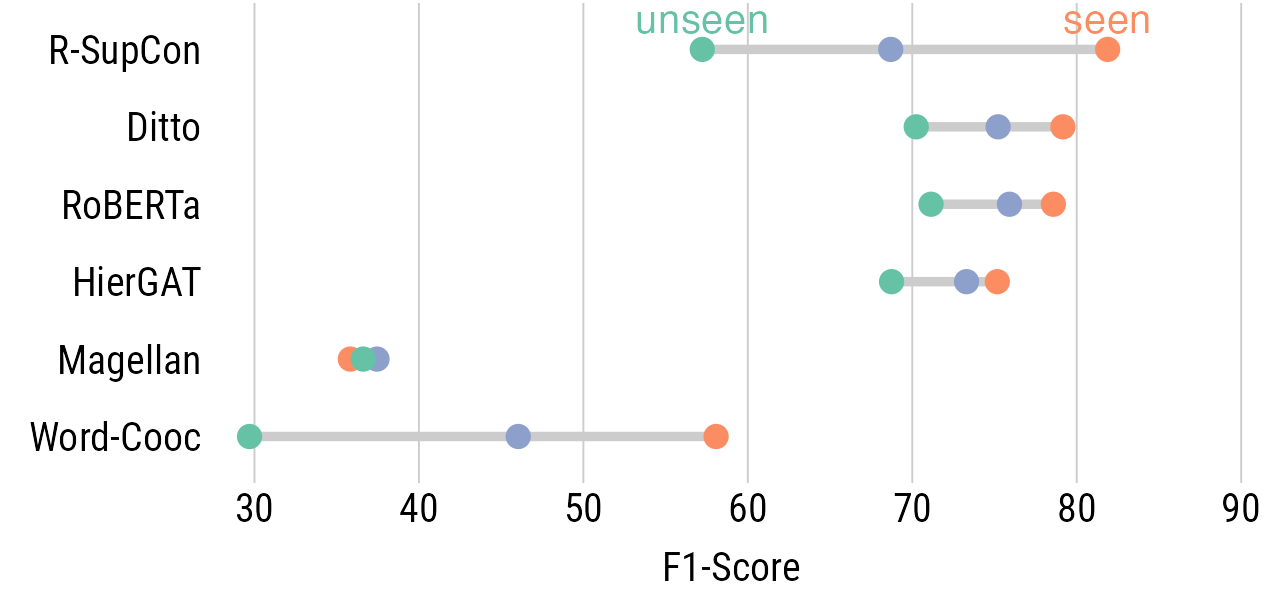}
  \caption{Varying fraction of unseen products in the test set combined with 50\% corner-cases and development set size medium.}
  \label{fig:unseen}
\end{figure}

\begin{figure}
  \centering
  \includegraphics[width=\columnwidth]{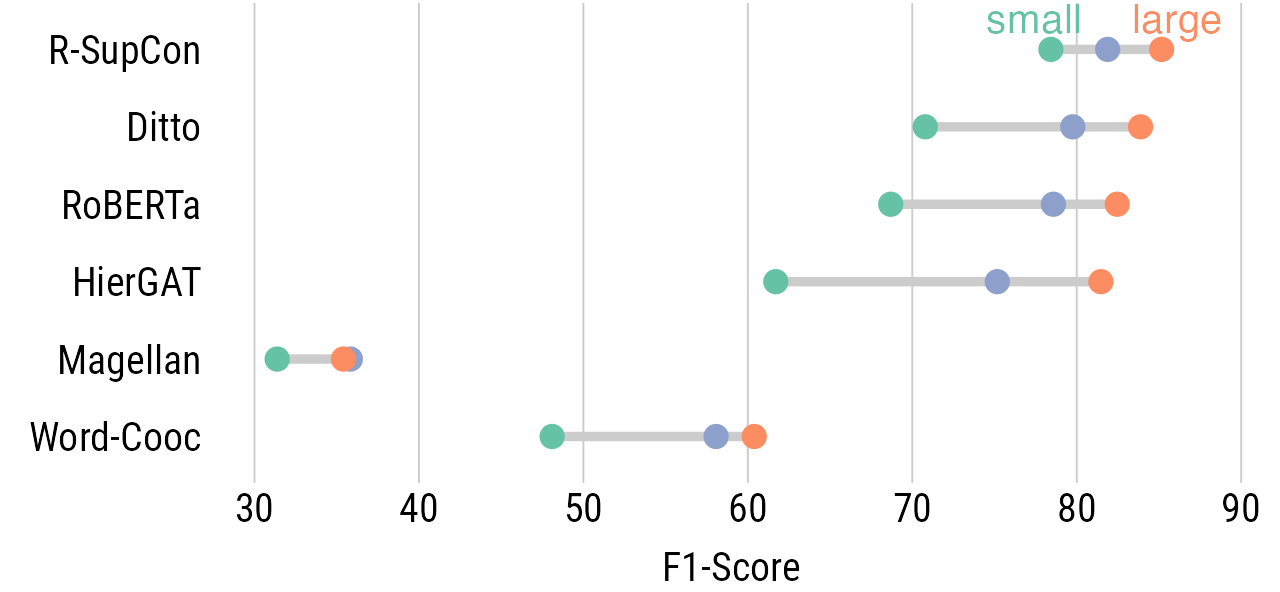}
  \caption{Varying development set sizes combined with 50\% corner-cases and no unseen entities in the test set.}
  \label{fig:training_size}
\end{figure}

Table \ref{tab:resultspair} summarizes the F1-score results of the experiments for the pair-wise matching tasks along all three dimensions: amount of corner-cases, amount of unseen entities in the test set, and development set size. Table \ref{tab:resultspair_pr} shows the corresponding precision and recall values for the four neural matching systems. Figures \ref{fig:hardness}, \ref{fig:unseen} and \ref{fig:training_size} visualize the results along the three dimensions.

\textbf{Dimension Corner-Cases:} Figure \ref{fig:hardness} fixes the development set size to \textit{medium} and unseen products to 0\% while varying the amount of corner-cases. Increasing the amount of corner-cases causes all methods to lose performance while not changing the ranking of the methods. The overall drop in performance is similar for all models suggesting that no model is much better suited for handling corner-cases. The absolute F1 values for the variants of the benchmark containing 80\% corner-cases for the top performing systems R-SupCon, Ditto, and RoBERTa all lie between 72.18 and 79.99, confirming that the WDC Products benchmark contains variants that are quite difficult for state-of-the-art systems. Varying the amount of corner cases negatively impacts the precision of matchers more than it does recall in general. This points to the matching systems mistaking very similar negatives as positives which is in line with the increased amount of these cases.

\textbf{Dimension Seen/Unseen:} Figure \ref{fig:unseen} shows the performance of all methods on test sets containing only seen, a mix of unseen and seen, as well as only unseen products while fixing the corner-case dimension to 50\% and development set size to \textit{medium}. It is clearly visible that all methods significantly drop in performance compared to completely seen data but specifically R-SupCon, which performs best on seen, experiences a large drop of 25\%. Nevertheless, all deep learning-based approaches significantly outperform the symbolic baselines even for unseen products. These same findings are visible on the remaining corner-case and development size sets. The unseen dimension negatively impacts the precision of all matchers significantly while recall generally increases, only R-SupCon experiences drops in performance on both measures, explaining the observed large drop in F1. The contrastive pre-training thus severely limits the generalization ability of this approach as the created clusters in the vector space are not useful for correctly disambiguating unseen products.

\textbf{Dimension Development Set Size:} Figure \ref{fig:training_size} shows the results for the development set size dimension while fixing the corner-cases to 50\% and unseen products to 0\%. The results show that most algorithms struggle with the \textit{small} development set apart from R-SupCon which can achieve an F1 of over 78\%. This gap closes significantly with the \textit{medium} development set for the deep learning methods. In general for pair-wise matching, even with the smallest development sets, the deep learning-based algorithms significantly outperform the symbolic baselines. With regards to the precision and recall measures, an increased size of the development set generally translates into improvements for both measures on all systems, although the increases in precision are generally more pronounced suggesting that the additional development data helps the models to better predict negative corner-cases.

\subsection{Multi-class Benchmark Results}
\label{subsec:resultsmulti}

Table \ref{tab:resultsmulti} shows the results of the multi-class matching experiment. R-SupCon exhibits a significantly higher performance than the other two baselines for multi-class matching. Interestingly, the simple symbolic word occurrence baseline is able to significantly beat the fine-tuned RoBERTa model for small and medium-size datasets for multi-class matching, an effect that is not observable for the pair-wise variant. The comparison of the multi-class RoBERTa model to its pairwise counterpart shows that both methods achieve similar results given the large development size but the multi-class model is not able to achieve the same results for the smaller sizes, suggesting, that for a multi-class formulation,
 fine-tuning a transformer model requires a minimum amount of three to four offers per class to achieve good results. Finally, R-SupCon reaches 3-6\% higher total F1s than its pair-wise counterpart even for smaller development set sizes, suggesting that this method is especially well suited for a multi-class matching scenario with the goal of recognizing a set of known products. The multi-class experiments underline the usefulness of the two versions for pair-wise and multi-class matching, as the performance of tested methods clearly differs between the versions.


\begin{table}[htb]
\centering
\caption{Results for multi-class matching over the dimensions development set size and amount of corner-cases. Results are micro-F1.}
\label{tab:resultsmulti}
\resizebox{\columnwidth}{!}{%
\begin{tabular}{@{}l|c|c|c|c@{}}
\toprule
\begin{tabular}[c]{@{}l@{}}Development \\ Set Size\end{tabular} & Corner-Cases          & Word-Occ & RoBERTa & R-SupCon \\ \midrule
Small                                                           & \multirow{3}{*}{80\%} & 63.30    & 36.63   & 82.30    \\
Medium                                                          &                       & 71.50    & 52.03   & 88.63    \\
Large                                                           &                       & 79.40    & 78.77   & 89.33    \\ \midrule
Small                                                           & \multirow{3}{*}{50\%} & 68.60    & 40.83   & 85.23    \\
Medium                                                          &                       & 76.10    & 61.33   & 89.80    \\
Large                                                           &                       & 81.10    & 82.00   & 91.73    \\ \midrule
Small                                                           & \multirow{3}{*}{20\%} & 66.60    & 39.83   & 87.87    \\
Medium                                                          &                       & 76.20    & 61.13   & 92.60    \\
Large                                                           &                       & 81.30    & 83.37   & 93.03    \\ \bottomrule
\end{tabular}%
}
\end{table}
\section{Related Work}
\label{sec:relatedwork}

\textbf{Entity Matching:}
Entity matching \cite{christophides_end--end_2020,BarlaugNeural2021,Christen2012DataMC,elmagarmidDuplicateRecordDetection2007} is the task of identifying entity descriptions that refer to the same real-world entity. Entity matching has been researched for over 50 years~\cite{fellegiTheoryRecordLinkage1969}. Early methods employ matching rules written by domain experts~\cite{fellegiTheoryRecordLinkage1969}. Further development led to better-performing methods based on symbolic unsupervised and supervised machine learning~\cite{christophides_end--end_2020}. 
Towards the end of the 2010s, the successes of deep learning algorithms in natural language processing and computer vision led to the development of the first deep learning-based methods for entity matching~\cite{mudgalDeepLearningEntity2018,shahNeuralNetworkBased2018}.
With the development of the Transformer~\cite{vaswaniAttentionAllYou2017} architecture and its implementations like BERT~\cite{devlinBERTPretrainingDeep2019} and RoBERTa~\cite{liu_roberta_2019} revolutionizing the performance in a wide range of natural language processing tasks, the data integration community has also turned to these language models for entity matching~\cite{brunnerEntityMatchingTransformer2020,liDeepEntityMatching2020,peetersDualobjectiveFinetuningBERT2021,yaoEntityResolutionHierarchical2022}.
More recent work started exploring the space of self-supervised and supervised contrastive losses~\cite{chenSimpleFrameworkContrastive2020,gaoSimCSESimpleContrastive2021,khoslaSupervisedContrastiveLearning2021} in combination with Transformer encoder networks for entity matching~\cite{peetersSupervisedContrastiveLearning2022a,wangSudowoodoContrastiveSelfsupervised2022}.
Other lines of  work consider graph-based methods~\cite{geCollaborERSelfsupervisedEntity2021,yaoEntityResolutionHierarchical2022} as well as domain adaption for entity matching~\cite{losterKnowledgeTransferEntity2021,trabelsiDAMEDomainAdaptation2022,tuDomainAdaptationDeep2022,akbarianrastaghiProbingRobustnessPretrained2022}.

\begin{table*}[htb]
\centering
\caption{Comparison of WDC Products to existing original benchmarks that have been used in recent entity matching work or have been recently proposed. See \cite{primpeliProfilingEntityMatching2020} for additional rarely used benchmarks. Splits marked with * have not been released with the respective benchmark but were adopted by the community. Dev and test size refer to the largest available split if applicable.}
\label{tab:comparison}
\resizebox{2.1\columnwidth}{!}{%
\begin{tabular}{@{}lccccccccccccc@{}}
\toprule
Benchmark &
  Domain &
  \# Sources &
  \# Entities &
  \# Records &
  \# Attr &
  \begin{tabular}[c]{@{}c@{}}Avg. \\ Density\end{tabular} &
  \# Matches &
  \begin{tabular}[c]{@{}c@{}}\# Non-\\ Matches\end{tabular} &
  \begin{tabular}[c]{@{}c@{}}Avg. Matches\\ per Entity\end{tabular} &
  \begin{tabular}[c]{@{}c@{}}Fixed Splits\\ (\# Train Sizes)\end{tabular} &
  \begin{tabular}[c]{@{}c@{}}Dev Size\\ (Matches)\end{tabular} &
  \begin{tabular}[c]{@{}c@{}}Test Size\\ (Matches)\end{tabular} &
  \begin{tabular}[c]{@{}c@{}}Max F1 in \\ related work\end{tabular} \\ \midrule
\multicolumn{14}{c}{Leipzig Database Group} \\ \midrule
Abt-Buy &
  Product &
  2 &
  1,012 &
  1,081/1,092 &
  3 &
  0.63 &
  1,095 &
  - &
  1.08 &
  \checkmark * (1) &
  7,659 (822) &
  1,916 (206) &
  94.29 \cite{peetersSupervisedContrastiveLearning2022a} \\
Amazon-Google &
  Product &
  2 &
  995 &
  1,363/3,226 &
  4 &
  0.75 &
  1,298 &
  - &
  1.30 &
  \checkmark * (1) &
  9,167 (933) &
  2,293 (234) &
  79.28 \cite{peetersSupervisedContrastiveLearning2022a} \\
DBLP-ACM &
  Bibliogr. &
  2 &
  2,220 &
  2,614/2,294 &
  4 &
  1 &
  2,223 &
  - &
  1.00 &
  \checkmark * (1) &
  9,890 (1,776) &
  2,473 (444) &
  99.17 \cite{liDeepEntityMatching2020} \\
DBLP-Scholar &
  Bibliogr. &
  2 &
  2,351 &
  2,616/64,263 &
  4 &
  0.81 &
  5,346 &
  - &
  2.27 &
  \checkmark * (1) &
  22,965 (4,277) &
  5,742 (1,070) &
  96.30 \cite{yaoEntityResolutionHierarchical2022} \\ \midrule
\multicolumn{14}{c}{DuDe} \\ \midrule
Restaurants &
  Company &
  2 &
  110 &
  533/331 &
  5 &
  1 &
  112 &
  - &
  1.02 &
  \checkmark * (1) &
  757 (88) &
  189 (22) &
  100.00 \cite{liDeepEntityMatching2020} \\
Cora &
  Bibliogr. &
  1 &
  118 &
  1,879 &
  18 &
  0.31 &
  64,578 &
  268,082 &
  547.27 &
   &
  - &
  - &
  100.00 \cite{wangEntityMatchingHow2011} \\ \midrule
\multicolumn{14}{c}{Magellan} \\ \midrule
Walmart-Amazon &
  Product &
  2 &
  846 &
  2,554/22,074 &
  10 &
  0.84 &
  1,154 &
  - &
  1.36 &
  \checkmark * (1) &
  8,193 (769) &
  2,049 (193) &
  88.20 \cite{yaoEntityResolutionHierarchical2022} \\
Company &
  Company &
  2 &
  28,200 &
  28,200/28,200 &
  1 &
  1 &
  28,200 &
  84,432 &
  1.00 &
  \checkmark * (1) &
  90,129 (22,560) &
  22,503 (5,640) &
  93.85 \cite{liDeepEntityMatching2020} \\
Beer &
  Product &
  2 &
  68 &
  4,345/3,000 &
  4 &
  0.96 &
  68 &
  382 &
  1.00 &
  \checkmark * (1) &
  359 (54) &
  91 (14) &
  94.37 \cite{liDeepEntityMatching2020} \\
iTunes-Amazon &
  Product &
  2 &
  120 &
  6,906/55,932 &
  7 &
  0.99 &
  132 &
  407 &
  1.10 &
  \checkmark * (1) &
  430 (105) &
  109 (27) &
  97.80 \cite{liDeepEntityMatching2020} \\ \midrule
\multicolumn{14}{c}{Alaska} \\ \midrule
Camera &
  Product &
  24 &
  103 &
  3,865 &
  56 &
  0.13 &
  157,157 &
  - &
  1,525.80 &
   &
  - &
  - &
  99.40 \cite{yaoEntityResolutionHierarchical2022} \\
Monitor &
  Product &
  26 &
  242 &
  2,283 &
  87 &
  0.17 &
  13,556 &
  - &
  56.02 &
   &
  - &
  - &
  99.60 \cite{yaoEntityResolutionHierarchical2022} \\ \midrule
\multicolumn{14}{c}{Chinese Academy of Sciences} \\ \midrule
Ember &
  Product &
  1 &
  350 &
  6,245 &
  5 &
  1 &
  5,053 &
  206,296 &
  14.44 &
  \checkmark (1) &
  8,000 (1,974) &
  50,000 (500) &
  78.45-96.89  \cite{wangBridgingGapReality2022a} \\ \midrule
\multicolumn{14}{c}{WDC} \\ \midrule
LSPM Computers &
  Product &
  269 &
  745 &
  3,665 &
  4 &
  0.51 &
  7,478 &
  59,571 &
  10.04 &
  \checkmark (4) &
  68,461 (9,690) &
  1,100 (300) &
  98.33 \cite{peetersSupervisedContrastiveLearning2022a} \\
LSPM Cameras &
  Product &
  190 &
  562 &
  4,068 &
  4 &
  0.43 &
  9,564 &
  35,899 &
  17.02 &
  \checkmark (4) &
  42,277 (7,178) &
  1,100 (300) &
  98.02 \cite{peetersDualobjectiveFinetuningBERT2021} \\
LSPM Watches &
  Product &
  235 &
  615 &
  4,676 &
  4 &
  0.5 &
  9,991 &
  53,105 &
  16.25 &
  \checkmark (4) &
  61,569 (9,264) &
  1,100 (300) &
  97.09 \cite{peetersDualobjectiveFinetuningBERT2021} \\
LSPM Shoes &
  Product &
  120 &
  562 &
  2,808 &
  4 &
  0.41 &
  4,440 &
  39,088 &
  7.90 &
  \checkmark (4) &
  42,429 (4,141) &
  1,100 (300) &
  97.88 \cite{peetersDualobjectiveFinetuningBERT2021} \\
\textbf{WDC Products} &
\textbf{  Product} &
  \textbf{3,259} &
  \textbf{2,162} &
  \textbf{11,715} &
  \textbf{5} &
  \textbf{0.79} &
  \textbf{28,299} &
  \textbf{124,899} &
  \textbf{13.09} &
  \textbf{\checkmark (3)} &
  \textbf{24,335 (8,971)} &
  \textbf{4,500 (500)} &
  \textbf{64.50-89.04} \\ \bottomrule
\end{tabular}%
}
\end{table*}

\textbf{Benchmarks for Entity Matching:} A sizable amount of publically accessible entity matching benchmarks exist today~\cite{primpeliProfilingEntityMatching2020}. 
In the following, we list a selection of the most important repositories for entity matching benchmarks sorted by time of creation. Table \ref{tab:comparison} summarizes key features of selected benchmarks.

The Database Group of the University of Leipzig\footnote{\url{https://dbs.uni-leipzig.de/research/projects/object\_matching/benchmark\_datasets\_for\_entity\_resolution}} published four benchmark tasks for pair-wise entity matching in 2010~\cite{kopckeEvaluationEntityResolution2010b} which are still widely used as of today~\cite{mudgalDeepLearningEntity2018,liDeepEntityMatching2020}. These benchmarks were released as two-source tasks where two tables and the perfect mapping between both  are available. The \textit{Abt-Buy} and \textit{Amazon-Google} benchmarks consist of mainly longer textual attributes from the product domain, while the \textit{DBLP-Scholar} and \textit{DBLP-ACM datasets} contain more structured less textual attributes.

The Hasso-Plattner Institute maintains the DuDe repository\footnote{\url{https://hpi.de/naumann/projects/data-integration-data-quality-and-data-cleansing/dude.html}} since 2011, providing three benchmark tasks~\cite{draisbachDuDeDuplicateDetection2017} originally published in other sources which have been modified for the entity matching setting with the \textit{Cora} benchmark considered more textual and the \textit{Restaurants} benchmark representing a very structured matching task. The benchmarks are provided together with the ground truth of matches.

The Magellan repository\footnote{\url{https://sites.google.com/site/anhaidgroup/useful-stuff/the-magellan-data-repository}} of the Data Management Research Group of the University of Wisconsin-Madison is maintained since 2015 and includes a large collection of benchmark tasks from other repositories as well as benchmarks created at the Research Group spanning a wide range of topics. The benchmarks span a wide range of topics and degrees of structuredness. The most prominent ones that have been used extensively in research are the \textit{Walmart-Amazon} (product), \textit{Company} (company), \textit{Beer} (product), and \textit{iTunes-Amazon} (product) benchmarks which apart from the highly textual Company, can all be considered structured benchmarks.

As all the previously presented benchmarks were created during a time were learning-based systems were not as prominent as they are today, no fixed training, validation, and test splits were originally made available with them. In 2018, the Deepmatcher~\cite{mudgalDeepLearningEntity2018} system, which was one of the first deep learning-based entity matching solutions, provided public downloads\footnote{\url{https://github.com/anhaidgroup/deepmatcher/blob/master/Datasets.md}} of the fixed splits for many of the benchmarks presented, which have consequently been adopted by the research community leading to better comparability of benchmark results across systems in recent works. 

The Alaska Benchmark\footnote{\url{http://alaska.inf.uniroma3.it/}}~\cite{crescenziAlaskaFlexibleBenchmark2021} of the Roma Tre University released in 2019 is a general benchmark for data integration tasks currently providing two e-commerce benchmarks for public download. The Alaska benchmarks support not only the entity matching task but further allow for an evaluation of schema matching systems due to the data being crawled from web pages of more than 20 different hosts which use different schemata for their product representations. Consequently, these benchmarks are highly structured to support the schema matching task. Labels are provided in the form of ground truth for matches. No negatives or fixed splits are available for these benchmarks yet but are supposed to be provided in the future.

The Web Data Commons (WDC) project maintained by our research group published the WDC Gold Standard for Product Matching and Product Feature Extraction\footnote{\url{http://webdatacommons.org/productcorpus/}} in 2016 and the WDC Product Data Corpus and Gold Standard for Large-Scale Product Matching (LSPM)\footnote{\url{http://webdatacommons.org/largescaleproductcorpus/v2/}} in 2019. Similar to the WDC Products benchmark, the LSPM benchmark is also based on schema.org product data which has been extracted from an earlier version of the Common Crawl, more specifically the November 2017 version.

Megagon Labs has published the \textit{Machamp} entity matching benchmark~\cite{wangMachampGeneralizedEntity2021} in 2021 which combines structured and unstructured tasks from previously described benchmark tasks into seven new tasks exhibiting differing degrees of structuredness. The \textit{Machamp} benchmark datasets provide fixed training, validation, and test splits together with the datasets to facilitate reproducibility and comparability.

In 2022, the Chinese-language product matching benchmark Ember~\cite{wangBridgingGapReality2022a} has been released. Similar to WDC Products, Ember also supports evaluating the generalization of matchers to unseen entities. Ember provides various test sets to investigate the impact of a higher negative-to-positive ratio among matches and non-matches as well as the utility of multi-modal features for product matching. Ember does not consider the development set size dimension and the amount of corner-cases. 

\textbf{Data Generators for Entity Matching:} In addition to the previously presented benchmarks, which rely on real-world datasets, a selection of data generators for entity matching exists \cite{christenFlexibleExtensibleGeneration2013,ioannouGeneratingBenchmarkData2013,ferraraBenchmarkingMatchingApplications2011,savetaLANCEPiercingHeart2015,hildebrandtLargeScaleDataPollution2020,primpeliImpactCharacteristicsMultisource2022} which allow for the creation of artificial datasets. All of the presented generators follow the general pattern of using a source dataset or repository of source entities which are then duplicated and corrupted using various transformation functions with varying degrees of corruption on the attributes of the data to generate heterogeneous matches.

\textbf{Benchmarks for Blocking:} The goal of blocking is to reduce the number of entity pairs that need to be processed by a matcher without losing recall~\cite{papadakisBlockingFilteringTechniques2020a}.   Many of the presented entity matching benchmarks are also used for evaluating blocking methods ~\cite{wangSudowoodoContrastiveSelfsupervised2022,brinkmannSCBlockSupervisedContrastive2023,papadakisBlockingFilteringTechniques2020a,papadakisComparativeAnalysisApproximate2016}. Recent work on blocking introduces larger benchmark datasets in order to challenge the scalability of blocking systems~\cite{papadakisComparativeAnalysisApproximate2016, papadakisHowReduceSearch2022}. Due to its size the PDC2020 product corpus (see Section \ref{subsec:extractionCC}) is well-suited as starting point for building blocking benchmarks. An example of a blocking benchmark that is derived from the corpus is the SC-Block benchmark~\cite{brinkmannSCBlockSupervisedContrastive2023}\footnote{\url{https://webdatacommons.org/largescaleproductcorpus/wdc-block/}}.

\textbf{Entity Matching as Non-Binary Classification:} While WDC Products provides two versions for pair-wise and multi-class entity matching, related work has nearly exclusively addressed entity matching as a binary pair-wise task. Multi-class classification methods are more prominent in the area of product categorization~\cite{sillaSurveyHierarchicalClassification2011,gaoDeepHierarchicalClassification2020} which constitutes a higher-level task aiming at groups of entities instead of single entities. Recent work introduced FlexER~\cite{genossarFlexERFlexibleEntity2022} which discusses the notion of what constitutes a match along a hierarchy of increasing strictness. Multi-class entity matching as presented in our benchmark is a relevant setup for practitioners as for instance similar multi-class experiments with proprietary data at eBay~\cite{shahNeuralNetworkBased2018} show. Additionally, recent work has tried to leverage multi-class matching to improve binary matching via multi-task learning~\cite{peetersDualobjectiveFinetuningBERT2021}. WDC Products tries to foster research on casting entity matching as multi-class classification by offering a multi-class formulation of the benchmark in addition to the pair-wise one.

\balance
\section{Adaption of Existing Benchmarks to the Dimensions of WDC Products} 
\label{sec:adaptionofbenchmarks}

This section discusses whether existing benchmarks already cover the dimensions of the WDC Products benchmark. It also explores the feasibility of extending existing benchmarks to implement these dimensions and points at  problems that hinder such an adaption.

\textbf{Development Set Size:} Most of the benchmarks listed in Table \ref{tab:comparison} only provide a single split and thus a fixed development set size (see column \textit{Fixed Splits} in Table \ref{tab:comparison}). The only benchmarks besides WDC Products that define development sets of different sizes are the LSPM benchmarks. It is of course possible to downsample the training and validation sets of all benchmarks in order to generate development sets of different sizes. A problem that arises from the downsampling is that most benchmarks aside from Alaska and WDC only contain a single match for most entities (see column \textit{Avg. Matches per Entity} in Table \ref{tab:comparison}). Down-sampling these benchmarks will increase the amount of entities that are not covered in the training set (unseen dimension) which will make the benchmarks more difficult and blurs the distinction between the two dimensions.

\textbf{Amount of Corner Cases:} None of the existing benchmarks implements the dimension amount of corner cases nor do they offer test sets having different levels of difficulty. The corner case dimension is realized in WDC Products by the careful selection of sets of 500 products from the PDC2020 corpus using the set of similarity metrics described in Section \ref{subsec:building}. 
The huge amount of products that is contained in PDC2020 ensures the availability of very similar but distinct products. 
Most existing benchmarks are not accompanied by a product corpus that can be used to select additional similar products. In order to include a corner case dimension into existing benchmarks, one would instead have to filter out some entities in order to create easier versions of the benchmark. The reduction of corner-cases would entail a reduction in the overall size of the benchmarks which would blur the distinction between the size and the corner case dimensions.

\textbf{Unseen Entities:} The dimension unseen entities is inseparably linked to the splitting of product offers into development and test set. For most existing benchmarks, namely the Magellan and Leipzig benchmarks, the community has adopted a set of splits that were created using random splitting\footnote{\url{https://github.com/anhaidgroup/deepmatcher/blob/master/Datasets.md}} of product pairs after blocking, which does not explicitly consider the unseen dimension. Given the low amount of average matches per entity in Table \ref{tab:comparison} for these benchmarks, it is possible that, by chance, all offers for a specific product end up only in the test set resulting in an unseen entity. The explicit modeling of this dimension requires knowledge of all matching entity descriptions in a dataset, or if not available, the reduction of the dataset to the set of known matches, which would severely reduce the size of a benchmark. Followed by a careful splitting procedure, it would be possible to explicitly model this dimension in the existing benchmarks, albeit on a much smaller scale than WDC Products. In addition to revising the splits of the benchmarks themselves, a notion similar to our unseen dimension can be realized by training on one benchmark and evaluating on another which contains the same type of entities, e.g. products. This approach has been explored in related work on transfer learning~\cite{trabelsiDAMEDomainAdaptation2022,tuDomainAdaptationDeep2022,akbarianrastaghiProbingRobustnessPretrained2022} for entity matching.


\textbf{Multi-Dimensionality:} In summary, while all dimensions can be realized to a certain extent in some of the existing benchmarks in isolation, one has to keep in mind that the dimensions do interact with each other. For example, classifying unseen entities in an environment of low corner-cases can be significantly easier than classifying unseen entities when many corner-cases are present as our results in Section \ref{sec:experiments} show. As we have discussed, realizing a single one of our chosen dimensions is often already prohibitive in existing benchmarks, realizing all of them in order to explore their interaction is practically infeasible for most benchmarks. WDC Products offers the possibility of exploring such nuances along its 27 variants and is further supported by the large PDC2020 corpus that can be used as starting point for extending the benchmark or exploring further dimensions.
\section{Conclusion}
\label{sec:conclusion}

WDC Products complements the field of benchmarking entity matching systems with a new multi-dimensional benchmark consisting of real-world product offers originating from thousands of e-shops. The benchmark provides 27 variants for the comparison of systems along the dimensions (i) amount of corner-cases (ii) generalization to unseen entities and (iii) development set size. In comparison to other benchmarks, WDC Products contains a larger amount of examples per entity which enables us to model the entity matching task as a binary pair-wise classification task on the one hand and on the other hand as a multi-class matching task while ensuring comparability between both setups. 
This makes WDC Products the first entity matching benchmark which provides for the fine-grained evaluation of matching systems along three dimensions, as well as the first benchmark to offer evaluation setups for both pair-wise and multi-class matching.
An evaluation of the utility of the benchmark using several recent matching systems shows that the benchmark is not easy to solve even for the most advanced systems. The evaluation also reveals that systems that exhibit a high performance on seen entities all struggle with the generalization to unseen entities. WDC Products is further accompanied by a large product corpus that can be used to extend the benchmark or explore additional dimensions.
We hope that the entity matching research community will perceive WDC Products as useful for the evaluation of their systems.

\begin{acks}
The authors acknowledge support by the state of Baden-Württemberg through bwHPC.
\end{acks}

\bibliographystyle{ACM-Reference-Format}
\bibliography{refs}

%
% ...

\end{document}